\documentclass[sigconf, screen, nonacm]{acmart}

\usepackage{graphicx}
\usepackage{subcaption}
\usepackage{booktabs}
\usepackage{algorithm}
\usepackage{algorithmic}

\usepackage{multirow}
\usepackage{colortbl}
\usepackage{xcolor}
\usepackage{dblfloatfix}
\usepackage{placeins}
\usepackage{footnote}
\AtBeginDocument{%
  }

\begin{document}

\title{Ride the Wave: Precision-Allocated Sparse Attention for Smooth Video Generation}


\author{Wentai Zhang}
\email{zhangwentai720@bupt.edu.cn}
\affiliation{%
  \institution{Beijing University of Posts and Telecommunications}
  \city{Beijing}
  \country{China}
}

\author{RongHui Xi}
\email{xironghui1@bupt.edu.cn}
\affiliation{%
  \institution{Beijing University of Posts and Telecommunications}
  \city{Beijing}
  \country{China}
}

\author{Shiyao Peng}
\email{pengshiyao@bupt.edu.cn}
\affiliation{%
  \institution{Beijing University of Posts and Telecommunications}
  \city{Beijing}
  \country{China}
}

\author{Jiayu Huang}
\email{jyxhuangjiayu@bupt.edu.cn}
\affiliation{%
  \institution{Beijing University of Posts and Telecommunications}
  \city{Beijing}
  \country{China}
}

\author{Haoran Luo}
\email{haoran.luo@ieee.org}
\affiliation{%
  \institution{Nanyang Technological University}
  \city{Singapore}
  \country{Singapore}
}

\author{Zichen Tang}
\email{tangzichen@bupt.edu.cn}
\affiliation{%
  \institution{Beijing University of Posts and Telecommunications}
  \city{Beijing}
  \country{China}
}

\author{HaiHong E}
\authornote{Corresponding author}
\email{ehaihong@bupt.edu.cn}
\affiliation{%
  \institution{Beijing University of Posts and Telecommunications}
  \city{Beijing}
  \country{China}
}


\begin{abstract}
  Video Diffusion Transformers have revolutionized high-fidelity video generation but suffer from the massive computational burden of self-attention. While sparse attention provides a promising acceleration solution, existing methods frequently provoke severe visual flickering caused by static sparsity patterns and deterministic block routing. To resolve these limitations, we propose \textbf{P}recision-\textbf{A}llocated \textbf{S}parse \textbf{A}ttention (PASA), a training-free framework designed for highly efficient and temporally smooth video generation. First, we implement a curvature-aware dynamic budgeting mechanism. By profiling the generation trajectory acceleration across timesteps, we elastically allocate the exact-computation budget to secure high-precision processing strictly during critical semantic transitions. Second, we replace global homogenizing estimations with hardware-aligned grouped approximations, successfully capturing fine-grained local variations while maintaining peak compute throughput. Finally, we incorporate a stochastic selection bias into the attention routing mechanism. This probabilistic approach softens rigid selection boundaries and eliminates selection oscillation, effectively eradicating the localized computational starvation that drives temporal flickering.
  Extensive evaluations on leading video diffusion models demonstrate that PASA achieves substantial inference acceleration while consistently producing remarkably fluid and structurally stable video sequences.
\end{abstract}


\begin{teaserfigure}
  \centering
  \includegraphics[width=0.49\textwidth]{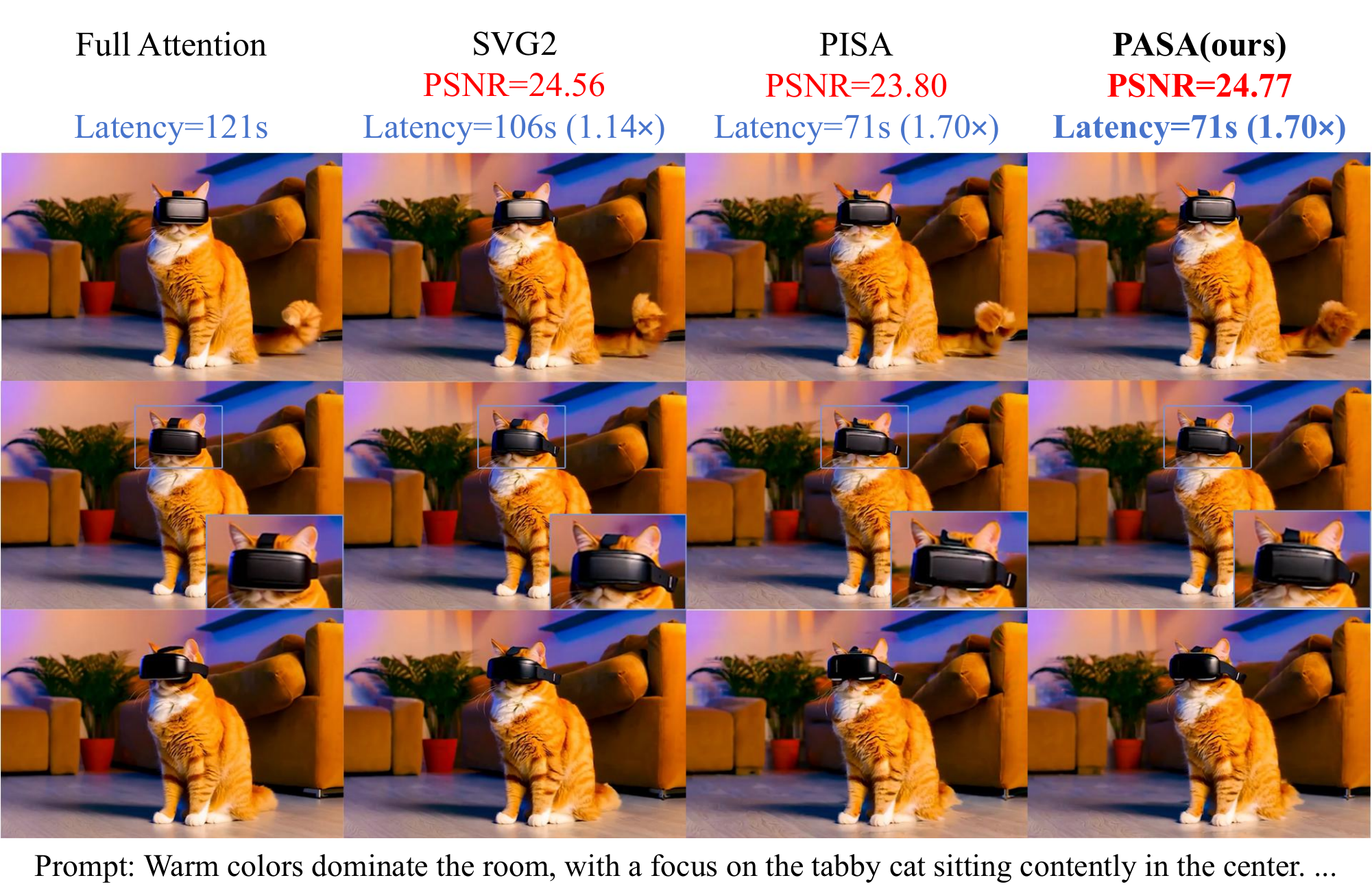}\hfill
  \includegraphics[width=0.49\textwidth]{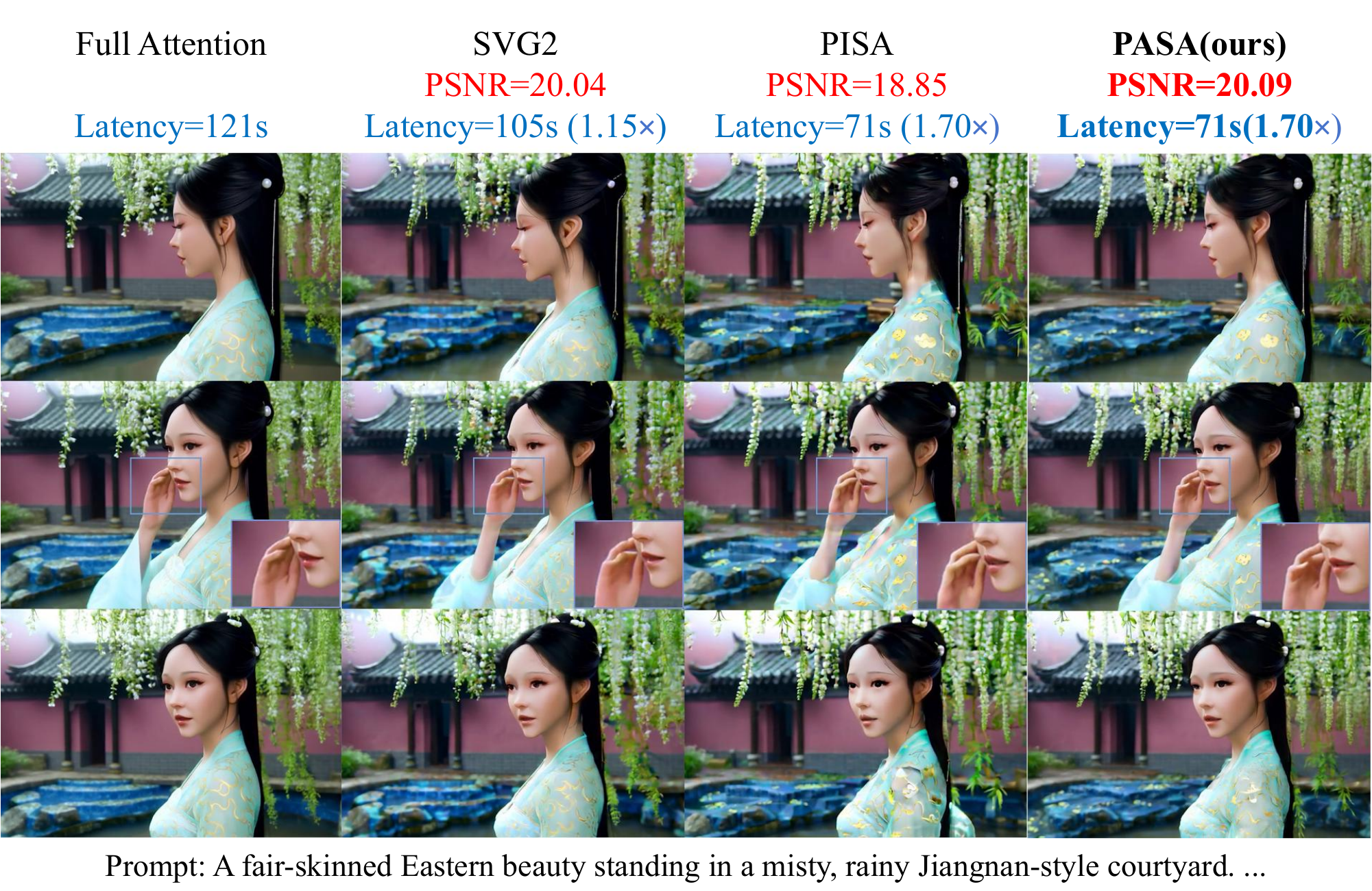}
  \caption{Visual and quantitative comparison on Wan~2.1-T2V-1.3B (quality \emph{vs.}\ efficiency). PASA (ours) matches the end-to-end speedup of PISA (\emph{Piecewise Sparse Attention}) while improving PSNR over other sparse attention baselines. \emph{Left:} PASA preserves continuous strap geometry; other sparse methods show discontinuities. \emph{Right:} PASA avoids the anatomical distortion present in competing sparse methods.}
  \label{fig:teaser}
\end{teaserfigure}

\received{20 February 2007}
\received[revised]{12 March 2009}
\received[accepted]{5 June 2009}

\maketitle

\section{Introduction}
\label{sec:intro}

With the rapid advancement of Artificial Intelligence Generated Content (AIGC) technologies, video generation, e.g., Text-to-Video (T2V)~\cite{t2v_hong2022cogvideo, t2v_villegas2022phenaki, t2v_wang2023modelscope} and Image-to-Video (I2V)~\cite{i2v_wang2023videocomposer, i2v_zhang2023i2vgen}, has emerged as a prominent research frontier, enabling automatic synthesis of realistic videos from textual descriptions or static images. The Diffusion Transformer (DiT)~\cite{ma2024latte, peebles2023scalable} architecture has demonstrated exceptional performance. However, for high-resolution or long-duration videos, the token sequence length ($n$) grows exceedingly large. A five-second 720p video comprises 75,600 tokens in Wan~2.1-14B~\cite{wan2025wan}. Since standard self-attention computes pairwise relationships across all tokens, it exhibits $O(n^2)$ computational and memory complexity and accounts for over 70\% of end-to-end latency~\cite{sparsevdit}, thus emerging as the primary bottleneck for efficient DiT inference.

To mitigate the quadratic computational complexity of attention in Diffusion Transformers, recent studies have actively explored sparse attention mechanisms. These existing approaches can be broadly categorized into two paradigms: selection-based and compensation-based methods. Selection-based methods~\cite{svg, svg2, spargeattention, sparsevdit} typically employ a binary ``keep-or-drop'' strategy. They dynamically identify critical token blocks to compute at a fine-grained level while entirely discarding the remaining non-critical blocks. Although hardware-friendly, this hard truncation inevitably leads to severe information loss and visual degradation under high sparsity. To alleviate this, recent compensation-based methods~\cite{rettention_2025, li2025psapyramidsparseattention, li2026pisapiecewisesparseattention, rectified_spaattn_2025} propose to preserve contextual information instead of outright discarding it. Rather than completely omitting non-critical blocks, these approaches utilize lightweight, coarse-grained computations—such as hierarchical pooling, statistical reshaping, or block-wise approximation—to compensate for the omitted attention weights.

\noindent\textbf{Limitation 1: Sub-optimal uniform sparsity across the denoising steps.}\quad Existing methods~\cite{sparsevdit, svg, spargeattention} typically apply a fixed top-$k$ or top-$p$ block budget uniformly across all inference timesteps. However, from the perspective of flow matching, the generation trajectory exhibits varying degrees of curvature and velocity acceleration. During highly non-linear phase transitions, the model requires dense contextual information to ensure semantic fidelity, whereas during stable, near-linear phases, a high computation budget is largely redundant. A static sparsity constraint fails to capture this temporal dynamic, leading to either wasted computation or severe quality degradation during critical timesteps.

\noindent\textbf{Limitation 2: Temporal flickering induced by deterministic block routing.}\quad Current dynamic sparse attention mechanisms rely on strictly deterministic scoring functions to select critical blocks. In the context of video generation, even microscopic variations in latent features across continuous frames can trigger abrupt, discontinuous shifts in the selected top-$k$ indices—a phenomenon we term "selection oscillation." This rigid, deterministic boundary causes sudden alterations between adjacent frames, which inevitably manifest as noticeable temporal flickering and jagged visual transitions, severely degrading the smoothness of the generated videos.

\noindent\textbf{Limitation 3: The dilemma between approximation granularity and hardware efficiency.}\quad To avoid the memory-bound bottleneck of computing per-block statistics, the compensation-based method \emph{Piecewise Sparse Attention} (PISA)~\cite{li2026pisapiecewisesparseattention} substitutes the exact block-wise first-order Taylor compensation term with a single global statistic $\overline{H}$. While this global substitution ensures high hardware throughput, it excessively homogenizes the non-critical regions, washing out the diverse local representations. Crucially, this global approach overlooks the inherent grouped execution patterns of modern GPU kernels, thereby missing a critical "sweet spot" that could balance local approximation fidelity with hardware efficiency.

To address these limitations, we present \textbf{PASA}, a training-free sparse attention mechanism that adaptively aligns computational resources with the inherent dynamics of video generation. Inspired by the continuous trajectory of flow matching, we introduce \textbf{Curvature-aware dynamic budgeting} to quantify per-timestep velocity acceleration via the L1 distance between adjacent noise predictions, and we reallocate the exact top-$k$ budget across timesteps. \textbf{Stochastic selection bias} injects controlled noise into block scoring to soften rigid deterministic routing, attenuating selection oscillation and temporal flicker across consecutive frames. \textbf{Grouped first-order approximation} aggregates per-block Taylor compensation statistics (the $H_j$ terms in Section~\ref{sec:pisa}) within coalesced memory-access groups, retaining finer locality than a single global substitute without reverting to costly per-block materialization.
Figure~\ref{fig:teaser} previews a representative quality--efficiency comparison on Wan~2.1-T2V-1.3B.

We summarize our contributions as follows:
\begin{itemize}
  \item We propose a novel temporal allocation strategy that adjusts the exact-computation budget (top-$k$) across varying inference timesteps. By utilizing the L1 distance of predicted velocity fields to profile the trajectory acceleration, our method precisely invests computational resources during critical semantic transitions while maximizing inference speed during stable phases.
  
  \item We introduce a grouped approximation scheme that calculates and shares Taylor expansion compensation statistics strictly within coalesced memory-access groups. This design successfully bridges the gap between memory-bound fine-grained computation and over-smoothed global compensation, delivering high local representational fidelity with optimal hardware throughput.
  
  \item We design a stochastic selection mechanism to eradicate the temporal flickering caused by rigid deterministic block routing. By injecting a controlled random bias into the attention scoring function, we soften the selection boundaries and effectively prevent selection oscillation, ensuring highly fluid and temporally consistent visual transitions across continuous video frames.

  \item Extensive experiments on leading video generation models, including Wan~2.1 and HunyuanVideo, validate the superiority of our method. Our approach achieves substantial inference speedups while establishing superior performance in temporal consistency and visual quality metrics on the VBench~\cite{huang2023vbench} benchmark. Ablation studies further demonstrate that each design choice contributes to video quality improvements.
\end{itemize}

\section{Related Work}

Existing sparse attention for video generation can be broadly bifurcated into two trajectories: training-free and training-based approaches.

\subsection{Training-free Sparse Attention}
Training-free methods accelerate inference by directly harnessing attention sparsity, ostensibly without incurring additional training overhead. Our work is in this scope. In practice, these approaches typically adopt a coarse-to-fine hierarchical pipeline: partitioning tokens into localized blocks, dynamically estimating block-wise importance, and subsequently restricting exact attention computation exclusively to the identified critical blocks \cite{svoo_diT_sparsity}. Within this paradigm, methods vary primarily in their mask generation strategies. Early explorations frequently employ pre-defined static sparse patterns. For instance, STA \cite{sta} introduces a hardware-friendly sliding-tile attention to supersede conventional global 3D attention, whereas Radial Attention \cite{radialattention} applies a prior-driven multi-band mask featuring radially decaying attention windows and temporal sampling frequencies. To enhance adaptability across diverse visual contents, subsequent research has pivoted towards dynamic sparse attention. SpargeAttn \cite{spargeattention} and DraftAttention \cite{draftattention} perform online block importance estimation by aggregating token activations, thereby bypassing low-relevance blocks. Similarly, XAttention \cite{xattention_2025} utilizes an antidiagonal-sum proxy coupled with dynamic programming for adaptive thresholding. To address the specific spatio-temporal redundancies in video sequences, AdaSpa \cite{adaspa} leverages cross-step invariance for precise online pattern searching, while \emph{Sparse VideoGen2} (SVG2)~\cite{svg2} employs K-means clustering to organize semantically cohesive tokens into contiguous memory layouts, thereby optimizing hardware utilization. Furthermore, to mitigate the attention distribution distortion that typically emerges at high sparsity regimes, methods like Rectified SpaAttn \cite{rectified_spaattn_2025} and Re-ttention \cite{rettention_2025} statistically reshape and reallocate pooled attention weights to correct systematic biases.

\subsection{Training-based Sparse Attention}
Conversely, training-based approaches inject sparsity constraints directly into the pre-training or fine-tuning phases. By explicitly optimizing the model weights to accommodate sparse attention layers, these methods unlock the potential for extreme sparsity levels \cite{sla_2025}. Initial explorations, such as VSA \cite{vsa_2025} and VMoBA \cite{vmoba}, attempt to incorporate fully learnable sparse attention mechanisms during training. More recently, hybrid architectures that seamlessly integrate sparse and linear attention have emerged as the prevailing standard. For example, SLA \cite{sla_2025} stratifies attention weights into critical, marginal, and negligible tiers. It allocates dense computation to critical weights, applies $\mathcal{O}(N)$ linear attention to marginal ones, and prunes the negligible connections entirely. Through lightweight fine-tuning, SLA drastically amplifies sparsity while preserving generation fidelity. Its successor, SLA2 \cite{sla2_2025}, further advances this by introducing a differentiable routing mechanism. Concurrently, to sustain end-to-end performance under ultra-sparse conditions, SpargeAttention2 \cite{spargeattn2_2026} proposes a hybrid top-$k$ and top-$p$ dynamic masking criterion. This is coupled with a velocity-level distillation fine-tuning objective, which utilizes a frozen full-attention model as a supervisory signal to robustly prevent behavioral drift during adaptation.

\section{Preliminaries}
\label{sec:preliminaries}
In this section, we formalize scaled dot-product attention in Video Diffusion Transformers and recall sparse attention via a per-query active key set.
We then summarize PISA~\cite{li2026pisapiecewisesparseattention}: exact attention on selected blocks plus Taylor-based correction on the remainder, which our method extends.

\begin{figure*}[!t]
  \centering
  \includegraphics[width=\textwidth]{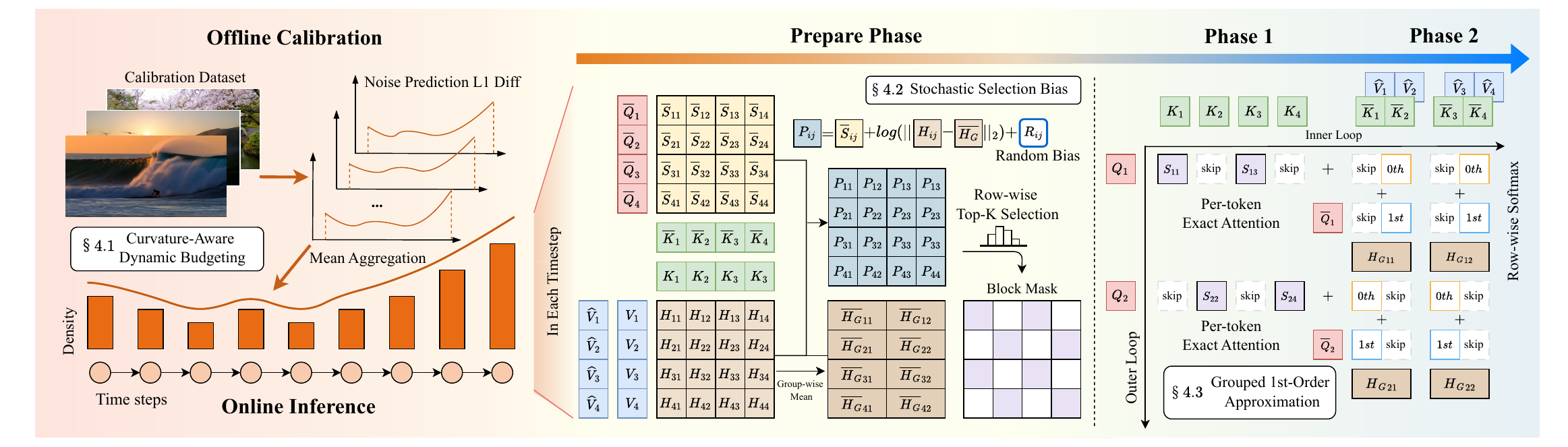}
  \caption{Overview of PASA: dynamic top-$k$ budgeting along the denoising trajectory, stochastic block scoring, and grouped first-order compensation for blocks in the unselected set.}
  \Description{Architecture diagram summarizing the PASA sparse attention pipeline for Video Diffusion Transformers.}
  \label{fig:overview}
\end{figure*}

\subsection{Self-Attention in Video Diffusion Transformers.}
In diffusion-based video generation with Transformers (Video DiTs), 
a video clip is first encoded into a compressed latent tensor using a pre-trained Variational Autoencoder (VAE). The input to a Transformer block is $X \in \mathbb{R}^{B \times N \times S \times D}$, with $B$ the batch size, $N$ the number of attention heads, $S$ the length of sequence, and $D$ the embedding dimension.
Given $X$, linear projections produce the Query ($Q$), Key ($K$),
and Value ($V$) matrices. With $N$ attention heads, each head attends over $S$ tokens; for clarity we omit batch size and head dimensions in the equations below.
The output $O$ is computed via scaled dot-product attention (with softmax applied row-wise):

\begin{equation}
O = \text{softmax}\!\left(\frac{Q K^T}{\sqrt{D}}\right) V
\end{equation}

The dominant computational bottleneck arises from the $Q K^\top$ matrix multiplication (forming the $S \times S$ attention logits) and the $P V$ matrix multiplication, where $P$ denotes the row-normalized attention weights from the softmax, both incurring $\mathcal{O}(S^2)$ time and memory complexity.

\subsection{Sparse Attention.}
Sparse attention avoids materializing full $S \times S$ interactions by selecting, for each query index $i \in \{1,\ldots,S\}$, a small active set of key indices $\mathcal{I}_i \subseteq \{1,\ldots,S\}$. A common construction assigns a routing score $r_{i,j}$ for each candidate key $j$ and takes the top-$k$ keys:
\begin{equation}
\mathcal{I}_i = \mathop{\mathrm{arg\,top}}_{j \in \{1, \dots, S\}}^k (r_{i,j})
\end{equation}

The attention output for query row $i$ uses only keys and values indexed by $\mathcal{I}_i$:
\begin{equation}
O_{i,:} = \text{softmax}\!\left(\frac{Q_{i,:} K_{\mathcal{I}_i}^\top}{\sqrt{D}}\right) V_{\mathcal{I}_i}
\end{equation}
where $Q_{i,:}$ is the $i$-th row of $Q$, and $K_{\mathcal{I}_i}$, $V_{\mathcal{I}_i}$ stack the corresponding rows of $K$ and $V$.

Keys outside $\mathcal{I}_i$ are typically skipped with coarse approximations. When each query attends to at most $k$ keys, cost drops from $\mathcal{O}(S^2)$ to $\mathcal{O}(S k)$ relative to full attention.

\subsection{Taylor Expansion Approximation for Sparse Attention}
\label{sec:pisa}
Instead of discarding unselected attention blocks as in conventional sparse attention, 
there is another approach~\cite{li2026pisapiecewisesparseattention} that approximates their contribution through a piecewise formulation that combines 
exact computation and analytical approximation.

Given query, key, and value matrices $Q,K,V \in \mathbb{R}^{L \times d}$, 
the sequence is partitioned into blocks of size $B$. 
For each query block, a scoring function estimates the importance of all key blocks. 
The top-$k$ blocks that form the selected set are computed exactly. The selected set is denoted as $\mathcal{S}$, and the remaining blocks form the unselected set $\mathcal{U}$. For the remaining blocks in $\mathcal{U}$, PISA approximates the attention contribution using a Taylor expansion around the block centroid. 
Let $\bar{K}_j$ denote the mean key vector of block $j$, and let 
$\alpha_{t,j} = \exp(Q_t \bar{K}_j^{\top})$ represent the approximate attention weight between query $Q_t$ and block $j$.
The zeroth-order approximation aggregates value vectors within the block:

\begin{equation}
\sum_{j \in \mathcal{U}} \alpha_{t,j} \left(\sum_{n=1}^{B} V_{j,n}\right)
\end{equation}

To further improve approximation accuracy, PISA introduces a first-order correction term based on the covariance between key and value vectors. 
Let

\begin{equation}
H_j = \sum_{n=1}^{B} (K_{j,n}-\bar{K}_j)^{\top} V_{j,n}
\end{equation}

Directly computing $H_j$ for every block results in a memory-bound operation. 
PISA therefore replaces block-wise statistics with a shared global statistic

\begin{equation}
\bar{H} = \frac{1}{N_B} \sum_{j=1}^{N_B} H_j
\end{equation}

where $N_B$ is the total number of blocks.
The final attention numerator combines exact sparse computation, block-wise zeroth-order approximation, and global first-order correction:

\begin{equation}
N_t =
\sum_{j \in \mathcal{S}} \sum_{n=1}^{B} \exp(Q_t K_{j,n}^{\top}) V_{j,n}
+
\sum_{j \in \mathcal{U}} \alpha_{t,j} \left(\sum_{n=1}^{B} V_{j,n}\right)
+
Q_t \bar{H} \sum_{j \in \mathcal{U}} \alpha_{t,j}
\end{equation}

This formulation preserves the contribution of long-tail attention blocks while avoiding quadratic complexity.
The approximation is integrated into the online softmax computation so that the normalization of attention weights remains consistent with full attention.
To determine the top-$k$ membership for the selected set $\mathcal{S}$, PISA employs a covariance-aware routing mechanism. Rather than evaluating block importance solely based on the centroid distance, the selection score integrates the semantic relevance with an approximation error prior, formulated as:
\begin{equation}
\text{Score}_{t,j} = \text{softmax}\!\left( \frac{Q_t \bar{K}_j^\top}{\sqrt{D}} + \log(||H_j - \bar{H}||_2 + \epsilon) \right)
\end{equation}
where $||H_j - \bar{H}||_2$ quantifies the structural heterogeneity of the key-value pairs within block $j$ based on the norm of its internal covariance matrix, and $\epsilon$ represents a small constant for numerical stability.
Our method is built on top of PISA, and we introduce the improvements in detail in Section \ref{sec:methodology}.

\section{Methodology}
\label{sec:methodology}

This section presents PASA (\textbf{P}recision-\textbf{A}llocated \textbf{S}parse \textbf{A}ttention), a training-free extension of PISA (Section~\ref{sec:preliminaries}). We depart from uniform per-step block budgets, deterministic top-$k$ routing, and a single global first-order correction for skipped blocks. Our method combines (\emph{i})~\emph{curvature-aware dynamic budgeting}, reallocating exact attention along the denoising trajectory in response to variability in the predicted velocity field; (\emph{ii})~\emph{stochastic routing}, perturbing block scores with lightweight randomness to mitigate persistent under-selection of background and boundary content and improve temporal stability; and (\emph{iii})~\emph{grouped first-order approximation}, sharing Taylor compensation statistics within memory-coalesced groups to preserve local fidelity while avoiding both global homogenization and fragmented per-block statistics. Figure~\ref{fig:overview} depicts the pipeline; the subsections below detail each module.

\subsection{Curvature-Aware Dynamic Budgeting}
\label{sec:curvature-aware-dynamic-budgeting}

\begin{figure*}[t]
  \centering
  \begin{subfigure}[b]{0.31\textwidth}
    \centering
    \includegraphics[width=\linewidth]{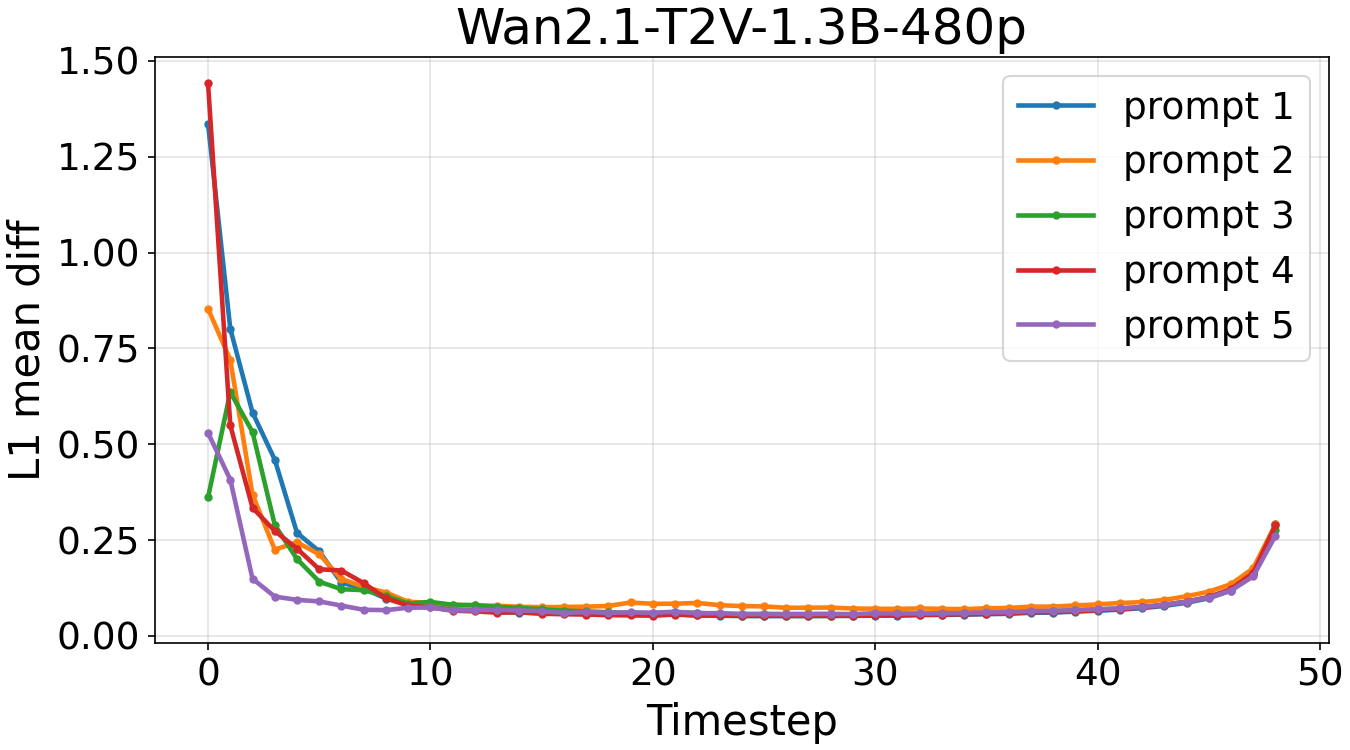}
    \caption{Wan~2.1-T2V-1.3B.}
    \label{fig:l1-mean-diff-wan13}
  \end{subfigure}\hfill
  \begin{subfigure}[b]{0.31\textwidth}
    \centering
    \includegraphics[width=\linewidth]{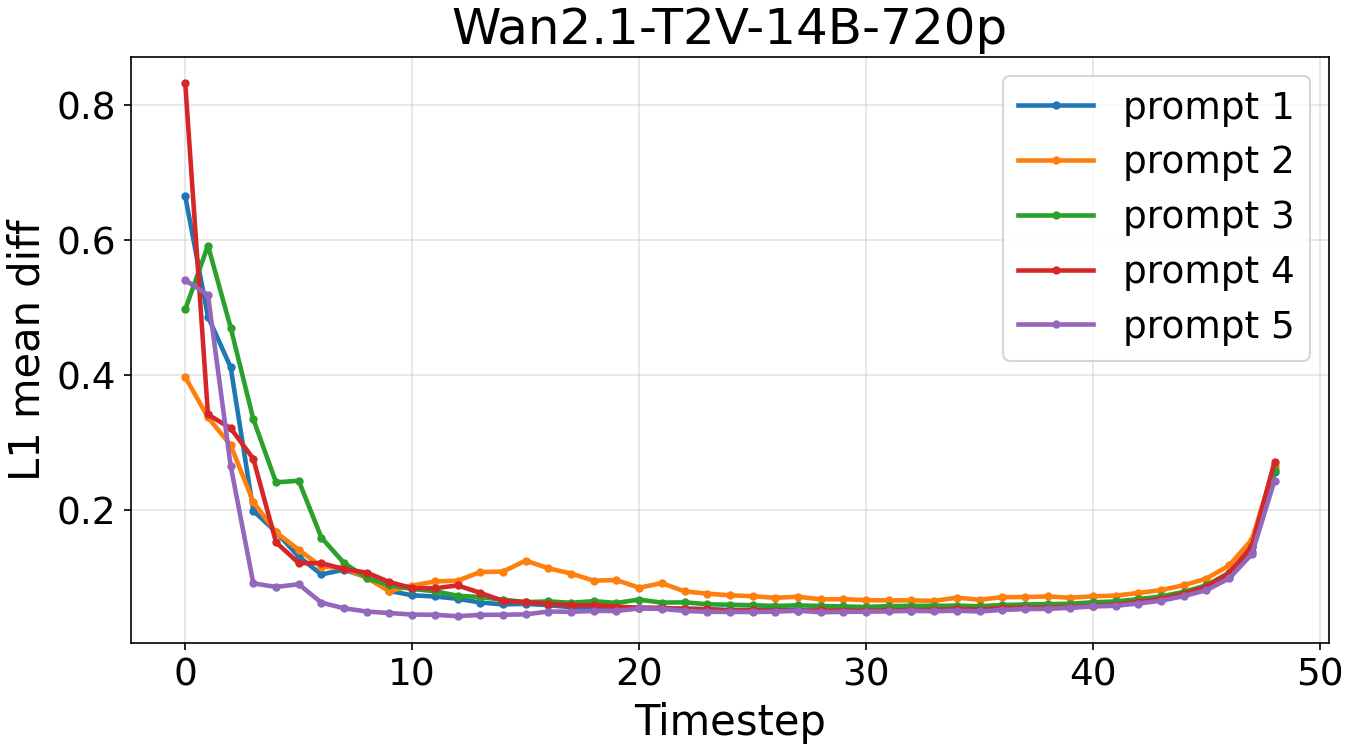}
    \caption{Wan~2.1-T2V-14B.}
    \label{fig:l1-mean-diff-wan14}
  \end{subfigure}\hfill
  \begin{subfigure}[b]{0.31\textwidth}
    \centering
    \includegraphics[width=\linewidth]{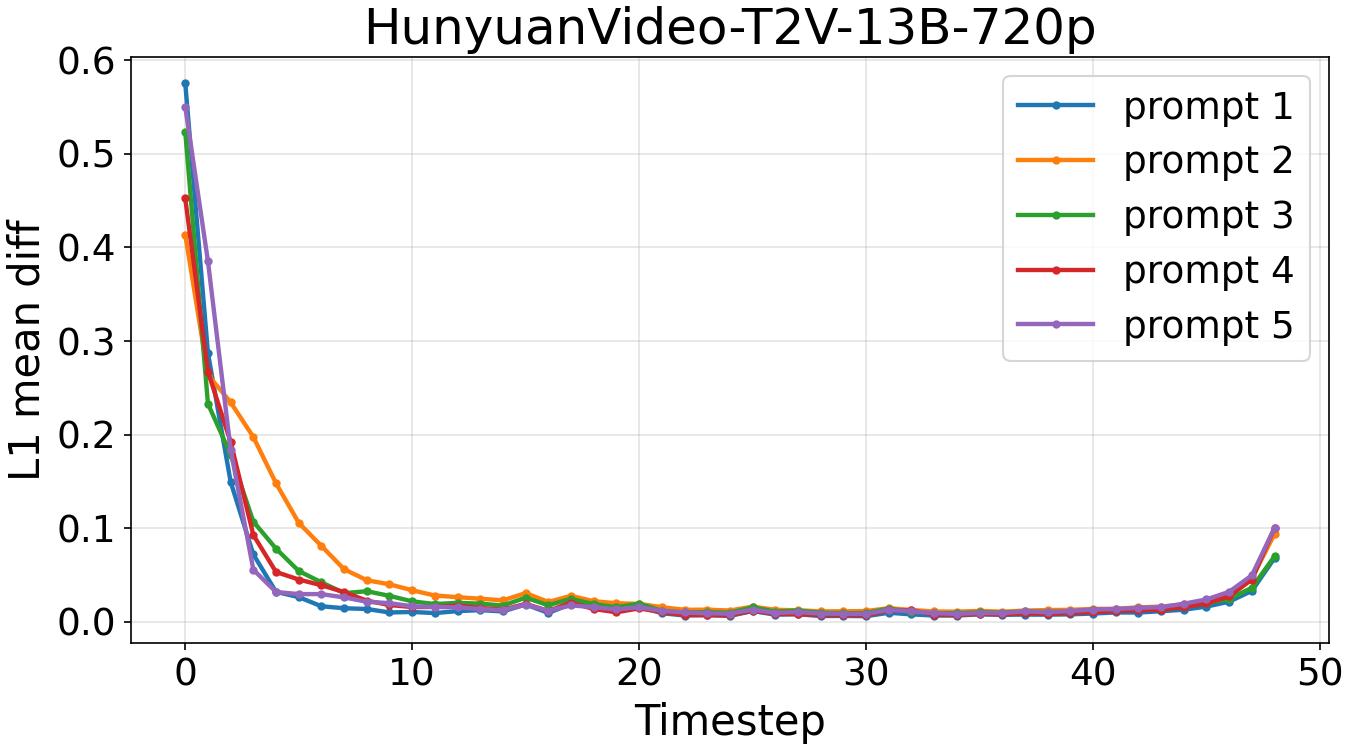}
    \caption{HunyuanVideo-T2V-13B.}
    \label{fig:l1-mean-diff-hyvideo}
  \end{subfigure}
  \caption{Mean L1 distance between predicted velocity fields at consecutive timesteps, averaged over prompts (per model).}
  \label{fig:l1-mean-diff-per-prompt}
\end{figure*}

Mainstream open-source video generation models, e.g., Wan~2.1 and HunyuanVideo, follow the flow-matching~\cite{lipman2023flowmatchinggenerativemodeling} formulation. The generation process dictates how a noise distribution transforms into the data distribution along a continuous probability flow trajectory. Our fundamental insight is that the generation trajectory does not evolve at a uniform pace. Instead, it exhibits varying curvature and velocity acceleration.
To motivate the design, we first visualize trajectory variability using five randomly selected prompts (separate from the calibration set below): for each prompt and model, we compute the L1 distance between predicted velocity fields~\footnote{The \texttt{noise\_pred} tensor in the model codebase.} at adjacent timesteps, shown in Figure~\ref{fig:l1-mean-diff-per-prompt}.

\emph{Insight.}
The generation process distinctly partitions into three phases. During the initial stage, which corresponds to approximately the first 10 timesteps, the magnitude of the velocity field variations is remarkably high and exhibits substantial variance across different prompts. This pronounced fluctuation indicates rapid, diverse macroscopic semantic construction. Consequently, this empirical observation provides a rigorous mathematical justification for prior studies, which typically mandate full dense attention computation during the initial 20\% to 25\% of the generation process to secure the global structural foundation.
Following this turbulent initial phase, the generation trajectory transitions into a prolonged stable regime. Throughout the intermediate timesteps, the rate of change in the velocity fields diminishes significantly. This consensus indicates that the probability flow follows a near-linear path, rendering a high computational budget largely redundant. Finally, as the generative process approaches the final 5 timesteps, the variation in the velocity fields experiences a distinct resurgence. Because each distinct prompt necessitates unique textural and spatial polishing, the generation trajectories correspondingly diverge into prompt-specific refinement directions. 

\emph{Solution.}
We use 10 calibration prompts (see Appendix~\ref{sec:appendix-calibration}) produced by a large language model. Following prior setups, we keep full attention on the first 20\% of denoising timesteps.
For the remaining 80\% timesteps $\mathcal{T}_{\mathrm{sparse}}$, we record the mean L1 distance between velocity fields at consecutive timesteps for each calibration prompt, yielding 10 curves; we average them into one curve that tracks global acceleration along the trajectory (combined with the budgeting pipeline in Figure~\ref{fig:overview}).
Let $\ell_t$ denote the prompt-averaged L1 signal at sparse timestep $t$.
PISA fixes a baseline top-$k$ \emph{density} $\rho$ (fraction of blocks computed exactly).
We summarize curvature along the sparse segment by the mean L1:
\begin{equation}
  \bar{\ell} = \frac{1}{|\mathcal{T}_{\mathrm{sparse}}|}\sum_{t \in \mathcal{T}_{\mathrm{sparse}}} \ell_t
\end{equation}
Mean-normalized scaling factors $\alpha_t$ with unit mean over $\mathcal{T}_{\mathrm{sparse}}$ are then
\begin{equation}
  \alpha_t = \frac{\ell_t}{\bar{\ell}}
\end{equation}
Finally, we multiply the baseline density to obtain the effective per-step density
\begin{equation}
  \rho_t = \rho \cdot \alpha_t
  \label{eq:dynamic-budget}
\end{equation}
Equation~\ref{eq:dynamic-budget} defines the effective per-step density and thus the sparse attention budget at $t$.
Because the $\alpha_t$ average to one over $\mathcal{T}_{\mathrm{sparse}}$, $\sum_{t \in \mathcal{T}_{\mathrm{sparse}}} \rho_t = \rho\,|\mathcal{T}_{\mathrm{sparse}}|$, matching the total budget of a uniform $\rho$; overall sparse attention compute is unchanged and only reallocated across timesteps.

\subsection{Stochastic Routing for Temporal Smoothness}

\emph{Insight.} We begin by investigating the root cause of temporal flickering in dynamic sparse attention. We hypothesize that compensation-based sparse methods partially mitigate video flickering specifically because they reduce the representational gap between the fully computed blocks and the approximated blocks. However, a severe structural imbalance persists. A standard video diffusion process typically passes the latent features through dozens of transformer layers across approximately fifty denoising timesteps. Under a strictly deterministic scoring mechanism, high-attention blocks, which predominantly capture moving subjects and salient foregrounds, are consistently selected for exact computation. Consequently, mid-attention and low-attention blocks, which typically comprise the background environment and the boundary regions between subjects, are persistently relegated to approximated computation. We hypothesize that this extremely skewed allocation of the computational budget between the primary subjects and the background context is a fundamental driver of temporal flickering, causing visual instability specifically within the background and along semantic boundaries.


\begin{figure}[t]
  \centering
  \includegraphics[width=\linewidth]{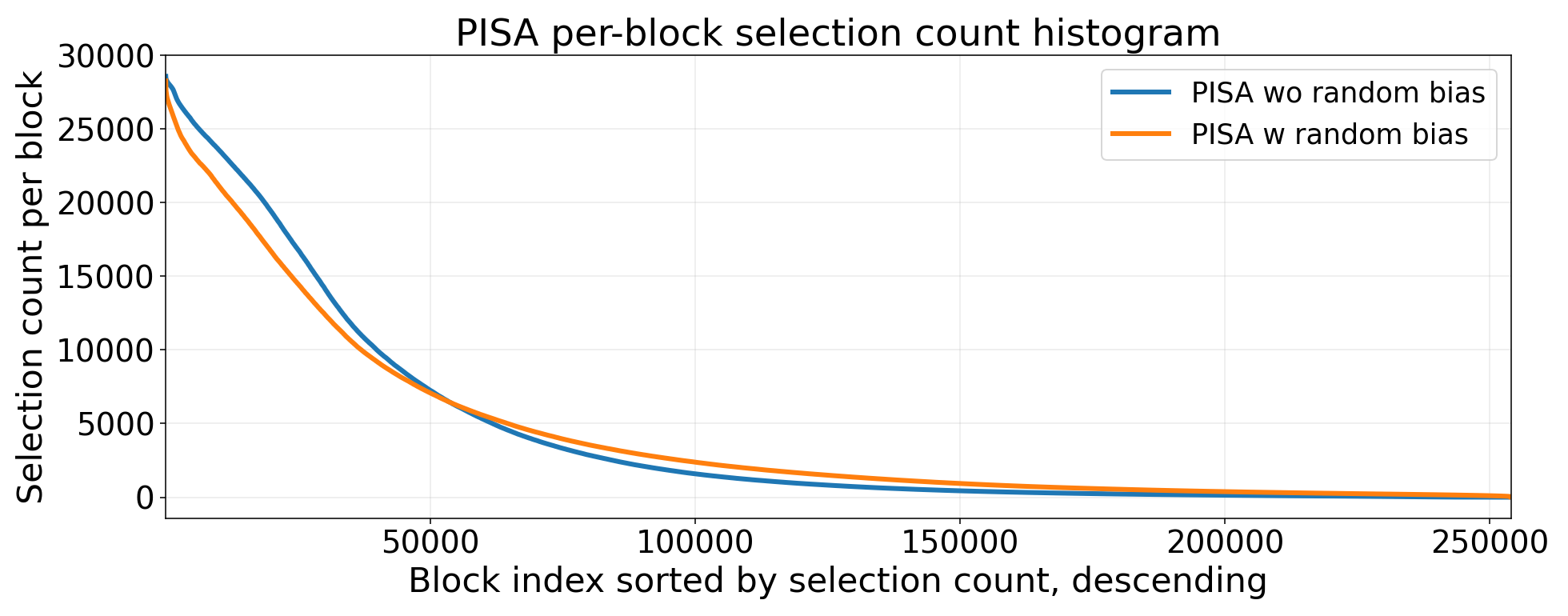}
  \caption{Distribution of exact-computation selection counts across all blocks during the complete inference process of the Wan~2.1-T2V-1.3B model. Introducing a stochastic selection bias effectively redistributes the computational budget from the over-selected primary blocks located in the left region to the under-selected intermediate and background blocks located in the middle and right regions.}
  \label{fig:selection_count}
\end{figure}

As visualized in Figure~\ref{fig:selection_count}, because the random bias is independently sampled across different layers and consecutive timesteps, the specific subset of mid-attention blocks upgraded to exact computation constantly shifts. This mechanism acts as a spatial-temporal multiplexer, ensuring that the entire background region receives periodic, fine-grained refinement rather than enduring continuous, coarse approximation. As demonstrated in our ablation studies (Section~\ref{sec:ablation_studies}), introducing this stochastic bias significantly improves objective metrics for both temporal flickering and motion smoothness.

\subsection{Grouped First-Order Approximation}
\emph{Insight.} Prior work compensates unselected attention blocks with a global first-order Taylor term $\bar{H}$ over the full unselected region, preserving subquadratic cost. This design is hardware-motivated: a single $\bar{H}$ (about 32\,KB) fits in fast on-chip SRAM on Hopper-class GPUs. The trade-off is representational: one global statistic homogenizes diverse non-critical content, blurring local background texture and fine spatial structure. Per-block exact statistics restore locality but trigger fragmented DRAM traffic that stalls the pipeline. We attribute the bottleneck less to storing localized statistics than to how those statistics are fetched---misaligned access patterns relative to the accelerator's asynchronous load-compute scheduling.

\emph{Solution.} We propose grouped first-order approximation: compute and share the Taylor compensation matrix within memory-coalesced groups of key and value tensors, avoiding both a lone global statistic and fully fragmented per-block updates. Triton-based operator profiling indicates that using 32 blocks per group better balances locality and throughput than coarser choices (e.g., 64). During the forward pass, our kernel streams each group's $\bar{H}$ from high-bandwidth memory together with the corresponding Key and Value tensors. We exploit warp specialization so compute warps execute attention arithmetic while load warps prefetch grouped statistics, overlapping fetch latency with computation. In practice, end-to-end latency matches the global baseline while improving attention fidelity; Appendix~\ref{sec:error-analysis} provides a formal error analysis with tightened bounds.

\section{Experiments}
\label{sec:experiments}

\subsection{Setup}

\subsubsection{Hardware and Software.}
All experiments were conducted on 8 NVIDIA H800 GPUs with CUDA 12.8. The Python version is 3.12.9. The Triton version is 3.4.0. 

\subsubsection{Models.} We conduct our experiments on three state-of-the-art open-weights Diffusion Transformer (DiT) models for text-to-video generation: \textbf{Wan~2.1-T2V-1.3B} and \textbf{Wan~2.1-T2V-14B}~\cite{wan2025wan} and \textbf{HunyuanVideo-T2V-13B}~\cite{kong2024hunyuanvideo}. Wan~2.1-T2V-1.3B is a leading DiT-based model with 1.3 billion parameters, utilizing a 3D-VAE for spatiotemporal compression to achieve high-fidelity motion dynamics. Wan~2.1-T2V-14B is a leading DiT-based model with 14 billion parameters, utilizing a 3D-VAE for spatiotemporal compression to achieve high-fidelity motion dynamics. HunyuanVideo-T2V-13B is another representative large-scale baseline. In Table~\ref{tab:main_result}, we use compact names (\emph{Hunyuan-13B}; Wan~2.1-1.3B / Wan~2.1-14B omitting the \emph{T2V} tag) to save column space. 

\subsubsection{Benchmarks.} To comprehensively evaluate video generation performance, we adopt VBench~\cite{huang2023vbench} to assess both visual quality and temporal consistency. Additionally, we employ SSIM, PSNR, and LPIPS~\cite{zhang2018unreasonable} to quantify the fidelity of our generated content compared to the full-attention baseline. Regarding efficiency, we prioritize practical end-to-end latency over purely theoretical throughput estimates, as the latter often diverge from real-world performance. Consequently, we report the actual end-to-end latency and the relative speedup ratio against the full attention implementation. Prompts for all experiments are from the Penguin Benchmark provided by the VBench team.

\subsubsection{Baseline.} 
We compare against \textbf{SVG2}~\cite{svg2} and \textbf{PISA}~\cite{li2026pisapiecewisesparseattention} (\emph{Sparse VideoGen2} and \emph{Piecewise Sparse Attention}), chosen as strong representatives of the two dominant sparse-attention paradigms.
Among selection-based (keep-or-drop) methods, SVG2 attains the best agreement with full-attention outputs under pixel-level fidelity metrics such as PSNR (and related similarity measures to dense-attention videos), making it the most competitive hard-pruning baseline. SVG2 improves block-sparse attention through a semantic-aware permutation strategy: tokens are first clustered via k-means and then permuted so that semantically similar tokens become contiguous in memory, enabling more accurate block-level attention approximation; cluster centroids are further used to estimate block importance and a top-$p$ policy selects the blocks for exact computation.
Among compensation-based methods, PISA currently yields the strongest VBench video-quality scores in this line of work; PASA extends PISA as described in Section~\ref{sec:pisa} and Section~\ref{sec:methodology}.

\subsection{Main Results.}

\begin{table*}[!t]
  \resizebox{\textwidth}{!}{
  \centering
  \small
  \setlength{\tabcolsep}{4pt}
  \begin{tabular}{l l c c c c c c c c c}
  \toprule
  \textbf{Model} & \textbf{Method} & \textbf{Sparsity} &
  \multicolumn{3}{c}{\textbf{VBench (\%) $\uparrow$}} &
  \multicolumn{3}{c}{\textbf{Similarity}} &
  \multicolumn{2}{c}{\textbf{Efficiency}} \\
  
   &  &  & \textbf{T.F.}& \textbf{M.S.}& \textbf{A.Q.}& \textbf{SSIM$\uparrow$} & \textbf{PSNR$\uparrow$} & \textbf{LPIPS$\downarrow$}
   & \textbf{Latency$\downarrow$} & \textbf{Speedup$\uparrow$} \\
  \midrule
  
  \multirow{4}{*}{\begin{tabular}{l}
  Wan~2.1-1.3B\\
  \textit{Text-to-Video}\\
  480P
  \end{tabular}}
  & Dense & 0.00\% & 97.36 & 98.27 & 66.34 & -- & -- & -- & 121 s & 1.00$\times$ \\
  & SVG2 & 84.5\% & 96.78 & 98.04 & \textbf{65.52} & \textbf{0.7765} & \textbf{21.52} & \textbf{0.1598} & \underline{86 s} & \underline{1.40$\times$} \\
  & PISA & 85\% & \underline{97.00} & \underline{98.08} & 65.17 & 0.7286 & 19.66 & 0.2246 & \textbf{71 s} & \textbf{1.70$\times$} \\
  & \cellcolor{gray!15}PASA (Ours)
  & \cellcolor{gray!15}85\%
  & \cellcolor{gray!15}\textbf{97.06}
  & \cellcolor{gray!15}\textbf{98.12}
  & \cellcolor{gray!15}\underline{65.33}
  & \cellcolor{gray!15}\underline{0.7510}
  & \cellcolor{gray!15}\underline{20.51}
  & \cellcolor{gray!15}\underline{0.2028}
  & \cellcolor{gray!15}\textbf{71 s}
  & \cellcolor{gray!15}\textbf{1.70$\times$} \\
  
  \midrule
  
  \multirow{4}{*}{\begin{tabular}{l}
  Wan~2.1-14B\\
  \textit{Text-to-Video}\\
  720P
  \end{tabular}}
  & Dense & 0.00\% & 98.47 & 98.95 & 65.08 & -- & -- & -- & 2786 s & 1.00$\times$ \\
  & SVG2 & 79.5\% & \underline{98.46} & \underline{98.92} & 64.17 & \textbf{0.8166} & \textbf{22.81} & \textbf{0.1345} & 1521 s & 1.83$\times$ \\
  & PISA & 85\% & 98.45 & \textbf{98.93} & \textbf{67.21} & 0.7794 & 21.20 & 0.1805 & \textbf{1252 s} & \textbf{2.23$\times$} \\
  & \cellcolor{gray!15}PASA (Ours)
  & \cellcolor{gray!15}85\%
  & \cellcolor{gray!15}\textbf{98.48}
  & \cellcolor{gray!15}\textbf{98.93}
  & \cellcolor{gray!15}\underline{64.83}
  & \cellcolor{gray!15}\underline{0.8003}
  & \cellcolor{gray!15}\underline{22.07}
  & \cellcolor{gray!15}\underline{0.1601}
  & \cellcolor{gray!15}\underline{1270 s}
  & \cellcolor{gray!15}\underline{2.19$\times$} \\
  
  \midrule
  
  \multirow{4}{*}{\begin{tabular}{l}
  Hunyuan-13B\\
  \textit{Text-to-Video}\\
  720P
  \end{tabular}}
  & Dense & 0.00\% & 98.99 & 99.25 & 62.29 & -- & -- & -- & 2415 s & 1.00$\times$ \\
  & SVG2 & 80.2\% & 99.03 & 99.24 & 62.45 & \textbf{0.8691} & \textbf{25.88} & \textbf{0.1039} & {1032 s} & {2.34$\times$} \\
  & PISA & 85\% & \underline{99.26} & \underline{99.38} & \underline{66.88} & 0.7907 & 23.74 & 0.1257 & \textbf{1004 s} & \textbf{2.40$\times$} \\
  & \cellcolor{gray!15}PASA (Ours)
  & \cellcolor{gray!15}85\%
  & \cellcolor{gray!15}\textbf{99.32}
  & \cellcolor{gray!15}\textbf{99.42}
  & \cellcolor{gray!15}\textbf{67.78}
  & \cellcolor{gray!15}\underline{0.8012}
  & \cellcolor{gray!15}\underline{25.24}
  & \cellcolor{gray!15}\underline{0.1168}
  & \cellcolor{gray!15}\underline{1020 s}
  & \cellcolor{gray!15}\underline{2.36$\times$} \\
  
  \bottomrule
  \end{tabular}}
  \caption{Quantitative comparison of our proposed Precision-Allocated Sparse Attention against state-of-the-art sparse methods across diverse video diffusion models. The abbreviated metrics represent Temporal Flickering (T.F.), Motion Smoothness (M.S.), and Aesthetic Quality (A.Q.). Bolded numerical values indicate the best results, while underlined values denote the second-best results.}
  \label{tab:main_result}
\end{table*}

\textbf{Overall Performance Trends.} As demonstrated in the main quantitative results in Table~\ref{tab:main_result}, our Precision-Allocated Sparse Attention consistently establishes the optimal equilibrium between computational efficiency and generation quality across all evaluated video diffusion models. Compared to the dense attention baseline, our approach achieves substantial inference acceleration, reaching speedups up to 2.36 times on the formidable Hunyuan-13B architecture. Furthermore, when compared directly to existing highly sparse mechanisms, our method universally outperforms the PISA baseline across all structural similarity and pixel-level distortion metrics. Simultaneously, our framework maintains an aggressive 85\% sparsity constraint, with end-to-end latency remaining close to the fastest sparse baselines per model (Table~\ref{tab:main_result}). This overarching trend confirms that our precision-allocation strategy successfully recovers critical visual details without sacrificing the raw hardware throughput gained from coarse mathematical approximations.

\textbf{Detailed Phenomenon Analysis.} Analyzing the fine-grained performance metrics reveals several distinct phenomena regarding temporal consistency and resource distribution. First, our method consistently achieves the highest scores in Temporal Flickering and Motion Smoothness among all sparse competitors, occasionally even surpassing the dense baseline on the Hunyuan-13B architecture. This pronounced temporal stability is a direct physical consequence of our stochastic routing mechanism, which guarantees periodic exact computation for intermediate background regions, thereby structurally eliminating the selection oscillation that plagues deterministic methods. Second, while SVG2 occasionally registers higher raw similarity scores on specific models, it relies on a significantly lower sparsity threshold, dynamically dropping to 79.5\% on the Wan~2.1-14B configuration. This lower sparsity directly translates to vastly inferior inference speedups. In stark contrast, our framework strictly enforces an 85 percent sparsity limitation yet effectively closes the visual fidelity gap. This phenomenon explicitly demonstrates that intelligently distributing a limited computational budget according to the generative trajectory curvature is fundamentally more effective than blindly retaining a larger volume of uniform attention blocks.

\subsection{Ablation Studies.}
\label{sec:ablation_studies}

\begin{figure}[t]
  \centering
  \begin{subfigure}[b]{\linewidth}
    \centering
    \includegraphics[width=\linewidth]{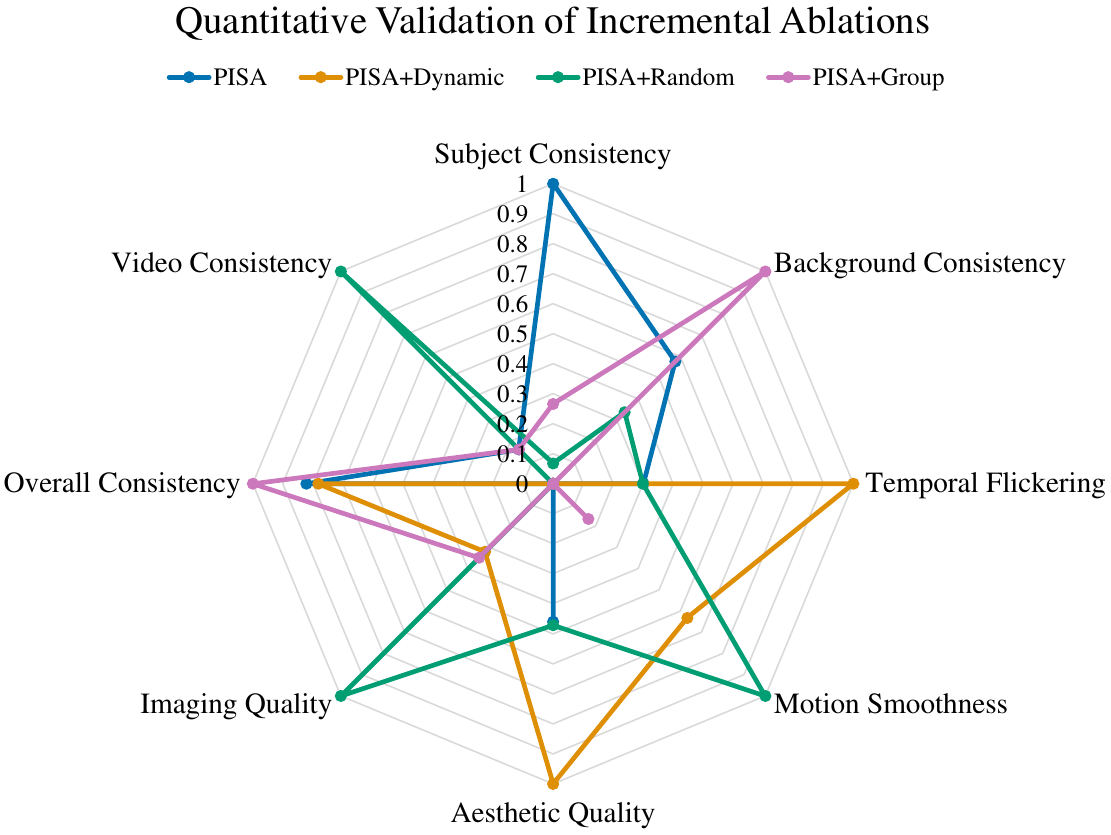}
    \label{fig:add-ablation}
  \end{subfigure}\hfill
  \begin{subfigure}[b]{\linewidth}
    \centering
    \includegraphics[width=\linewidth]{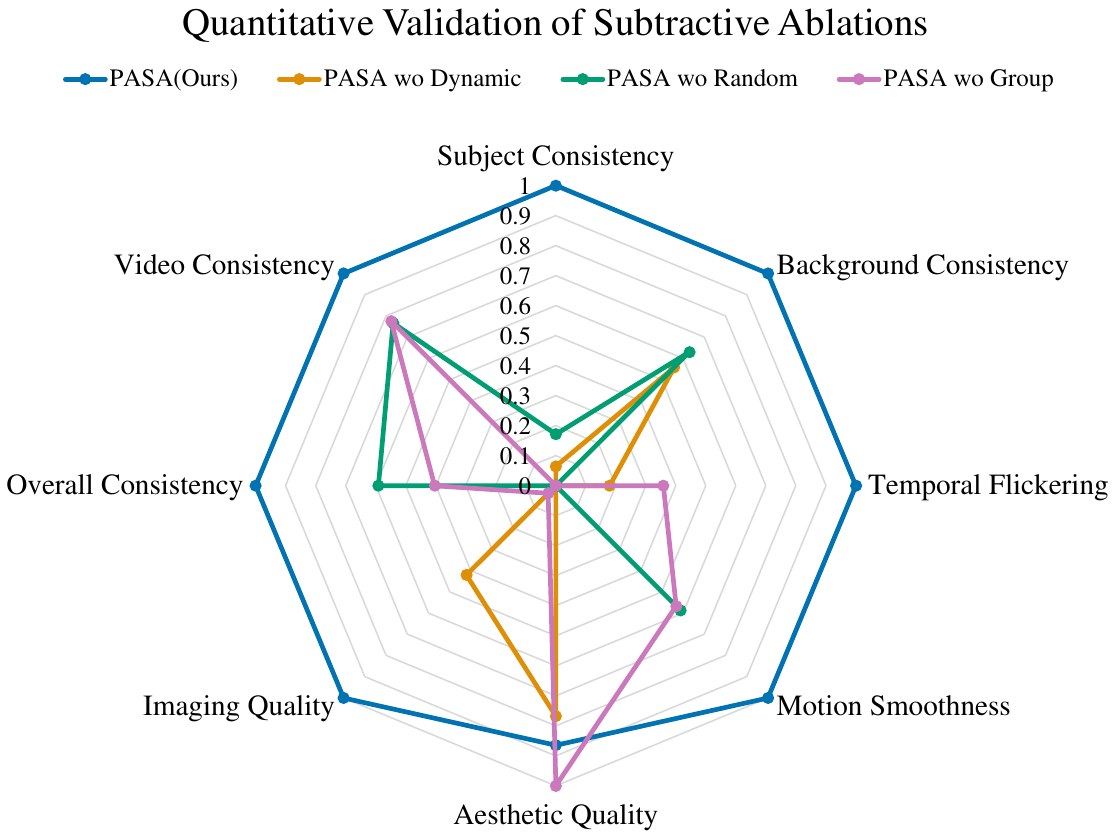}
    \label{fig:sub-ablation}
  \end{subfigure}
  \caption{Quantitative validation of incremental and subtractive ablation studies evaluated on the Wan~2.1-T2V-1.3B model. The performance is assessed across eight distinct dimensions of video generation quality. Notably, Video Consistency explicitly represents the PSNR. To facilitate a clear and unified comparative visualization across disparate measurement scales, all raw metric scores are normalized using min--max scaling. In the legend, ``Dynamic'' denotes the curvature-aware dynamic budgeting; ``Random'' signifies the stochastic selection bias; and ``Group'' represents the grouped first-order approximation.}
  \label{fig:ablation-add-sub}
\end{figure}

To rigorously evaluate the individual contributions and the synergistic effects of our proposed modules, we conduct comprehensive incremental and subtractive ablation studies utilizing the Wan~2.1-T2V-1.3B model, summarized in Figure~\ref{fig:ablation-add-sub}. The incremental study originates from the PISA baseline, progressively integrating our modules to isolate their additive benefits. Conversely, the subtractive study begins with our complete PASA, systematically removing individual components to expose structural vulnerabilities. We measure the generated videos across eight multidimensional criteria: Subject Consistency, Background Consistency, Temporal Flickering, Motion Smoothness, Aesthetic Quality, Imaging Quality, Overall Consistency, and Video Consistency (quantified via the PSNR). To construct the comparative radar charts, we apply min--max normalization to the raw numerical outcomes of each metric, anchoring the minimum observed value at zero and the maximum at one.
The resulting radar charts reveal several highly pronounced experimental phenomena, which we systematically analyze and explain below:

\textit{Observation 1: Grouped Approximation Rescues Background Fidelity.} 
In the incremental ablation chart, integrating the grouped first-order approximation into the baseline dramatically elevates Background Consistency and Overall Consistency to the maximum normalized score of one. Conversely, the subtractive chart reveals that removing the grouped first-order approximation from our complete framework causes Background Consistency to plummet directly to zero. 
\textbf{Explanation:} The baseline algorithm relies on a single sequence-wide global statistic that entirely homogenizes the non-critical regions. This mathematical over-smoothing effectively washes out localized environmental textures. By calculating the Taylor expansion compensation strictly within memory-coalesced groups, our hardware-aligned grouped approximation preserves fine-grained local variations. This directly validates our theoretical claim that localized group statistics prevent the severe textural degradation of the background environment.

\textit{Observation 2: Stochastic Routing Dictates Motion Fluidity.} 
The incremental addition of the random bias provides a massive enhancement to Motion Smoothness and Video Consistency. Corroborating this, the subtractive study demonstrates that eliminating the random bias heavily degrades Motion Smoothness and violently exacerbates Temporal Flickering.
\textbf{Explanation:} Deterministic routing mechanisms force the attention operation to abruptly toggle intermediate-attention blocks on and off across continuous temporal frames due to microscopic latent feature fluctuations. This selection oscillation severely disrupts temporal coherence. The injected stochastic bias functions as a probabilistic temporal low-pass filter, granting mid-attention blocks periodic and spatially shifting exact-computation refinement. This effectively resolves the localized computational starvation, translating directly into the observed fluidity in motion and the suppression of jarring structural shifts.

\textit{Observation 3: Dynamic Budgeting Governs Structural Stability.} 
The charts indicate that applying the dynamic budgeting comprehensively improves Temporal Flickering and Aesthetic Quality in the incremental study. Furthermore, removing it from the full framework severely cripples Subject Consistency and Temporal Flickering.
\textbf{Explanation:} The continuous probability flow trajectory involves distinct phases of rapid semantic transitions and stable linear evolution. A static computation budget dangerously under-allocates resources during critical high-curvature transitions. Our curvature-aware dynamic budgeting precisely identifies these acceleration phases utilizing the absolute distance of predicted velocity fields, strategically investing heavy computation exactly when the global structural foundation is being constructed. This dynamic alignment prevents the degradation of primary subjects and ensures robust aesthetic formulation throughout the turbulent stages of the generation trajectory.

\textit{Observation 4: The Necessity of Synergistic Integration.} 
The subtractive ablation radar chart clearly illustrates that our complete Precision-Allocated Sparse Attention framework universally dominates the outer perimeter across nearly all evaluation dimensions. 
\textbf{Explanation:} The three proposed mechanisms address fundamentally orthogonal limitations of sparse attention optimization. The grouped approximation restores spatial details, the stochastic routing ensures temporal fluidity, and the dynamic budgeting aligns resource allocation with the physical generation trajectory. Removing any single component creates a critical structural vulnerability that cannot be compensated for by the remaining modules. This comprehensive dominance proves that state-of-the-art generation quality relies strictly on their highly synergistic integration.

\section{Conclusion}
This work focus on the tension between sparse attention and temporal stability in Video Diffusion Transformers by introducing Precision-Allocated Sparse Attention (PASA), a training-free framework that includes curvature-aware dynamic budgeting, hardware-aligned grouped approximations, and stochastic attention routing. Together, these components concentrate exact computation on semantically critical trajectory segments, preserve fine-grained spatial detail without sacrificing throughput, and suppress the selection oscillation that causes flicker. Experiments on leading text-to-video models show that PASA delivers substantial speedups at high sparsity while improving or matching prior sparse methods on temporal consistency, motion smoothness, and overall video quality.


\clearpage

\bibliographystyle{ACM-Reference-Format}
\bibliography{sample-base}

\clearpage
\appendix

\section{Calibration Prompts}
\label{sec:appendix-calibration}
\noindent We used the GPT 5.3 LLM to generate 10 diverse text-to-video strings used solely to calibrate the budgeting curve in Section~\ref{sec:curvature-aware-dynamic-budgeting}.
\begin{itemize}
  \item A golden sunrise over a quiet mountain lake, mist floating above the water, birds flying across the sky, cinematic lighting, ultra-realistic, gentle camera pan, 4K.
  \item A bustling street café in Paris on a sunny morning, people chatting, waiters serving coffee and croissants, light breeze moving umbrellas, warm cinematic atmosphere.
  \item A small robot watering plants in a cozy greenhouse, sunlight streaming through glass panels, colorful flowers everywhere, soft depth of field, whimsical mood.
  \item A surfer riding a large ocean wave at sunset, glowing orange sky reflecting on the water, slow motion splash, cinematic wide shot, dramatic lighting.
  \item A peaceful Japanese garden in spring, cherry blossoms falling slowly, koi fish swimming in a clear pond, stone lanterns and wooden bridge, tranquil atmosphere.
  \item A futuristic city skyline at night, flying cars moving between glowing skyscrapers, neon reflections on wet streets, cyberpunk aesthetic, smooth aerial camera movement.
  \item A group of friends hiking through a lush forest trail, sunlight filtering through tall trees, laughter and conversation, natural documentary style.
  \item A cozy winter cabin in the mountains, snow gently falling outside, warm fireplace glowing inside, steam rising from a cup of hot chocolate, cinematic interior shot.
  \item A colorful hot air balloon festival at sunrise, dozens of balloons slowly lifting into the sky above rolling hills, soft morning light, drone shot.
  \item A playful golden retriever running across a green field, chasing a red ball, bright blue sky with fluffy clouds, joyful and energetic slow motion.
\end{itemize}

\section{Error Analysis}
\label{sec:error-analysis}

In this section, we relate the grouped first-order surrogate in PASA to the global $\bar{H}$ used in PISA (Section~\ref{sec:pisa}). We separate a standard \emph{within-group} optimality statement (sum of squared matrix deviations) from a \emph{weighted residual} bound; we do \emph{not} claim a pointwise ordering of per-block deviations from group versus global means, which would be false in general.

\textbf{Mathematical Formulation and Notation.} Let $H_j \in \mathbb{R}^{d \times d}$ denote the first-order Taylor statistic for block $j$ as in Section~\ref{sec:pisa}. The PISA global surrogate averages over all $N_B$ blocks: $\bar{H}_{\text{global}} = \frac{1}{N_B} \sum_{j=1}^{N_B} H_j$, matching $\bar{H}$ there.

We partition the $N_B$ blocks into $K$ disjoint hardware groups $\mathcal{G}_1, \ldots, \mathcal{G}_K$ and set
\begin{equation}
\bar{H}_{\text{group}}^{(g)} = \frac{1}{|\mathcal{G}_g|} \sum_{j \in \mathcal{G}_g} H_j
\end{equation}

Fix a query row $q_t$. Let $\mathcal{U}_t$ be the set of \emph{unselected} blocks for that query, $\mathcal{U}_{t}^{(g)} = \mathcal{U}_t \cap \mathcal{G}_g$, and let $\alpha_{t,j} = \exp(Q_t \bar{K}_j^{\top})$ denote the same block-centroid weights as in Section~\ref{sec:pisa} (pre-softmax). The exact first-order numerator contribution from unselected blocks is
\begin{equation}
\mathcal{N}_{\text{exact}}^{(1)} = q_t \sum_{g=1}^K \sum_{j \in \mathcal{U}_{t}^{(g)}} \alpha_{t,j} H_j
\end{equation}
and the grouped surrogate uses $\bar{H}_{\text{group}}^{(g)}$ inside each group:
\begin{equation}
\tilde{\mathcal{N}}_{\text{group}}^{(1)} = q_t \sum_{g=1}^K \left( \sum_{j \in \mathcal{U}_{t}^{(g)}} \alpha_{t,j} \right) \bar{H}_{\text{group}}^{(g)}
\end{equation}

\textbf{Lemma 1 (Weighted group residual).} Define the matrix residual
\begin{equation}
R_{t,\text{group}} = \sum_{g=1}^K \sum_{j \in \mathcal{U}_{t}^{(g)}} \alpha_{t,j} \bigl(H_j - \bar{H}_{\text{group}}^{(g)}\bigr)
\end{equation}
Let
\begin{equation}
M_{\text{group}} := \max_{g}\ \max_{j \in \mathcal{U}_{t}^{(g)}} \bigl\| H_j - \bar{H}_{\text{group}}^{(g)} \bigr\|_{\mathrm{F}}
\end{equation}
where $\|\cdot\|_{\mathrm{F}}$ is the Frobenius norm. By the triangle inequality for $\|\cdot\|_{\mathrm{F}}$,
\begin{equation}
\| R_{t,\text{group}} \|_{\mathrm{F}} \;\le\; M_{\text{group}} \sum_{j \in \mathcal{U}_t} \alpha_{t,j}
\end{equation}

\textbf{Proposition 1 (Within-group sum of squares).} For each group $g$, the group mean $\bar{H}_{\text{group}}^{(g)}$ minimizes the aggregate squared Frobenius deviation over blocks in that group: for any matrix $C$,
\begin{equation}
\sum_{j \in \mathcal{G}_g} \bigl\| H_j - \bar{H}_{\text{group}}^{(g)} \bigr\|_{\mathrm{F}}^2 \;\le\; \sum_{j \in \mathcal{G}_g} \bigl\| H_j - C \bigr\|_{\mathrm{F}}^2
\end{equation}
In particular, with $C = \bar{H}_{\text{global}}$,
\begin{equation}
\sum_{j \in \mathcal{G}_g} \bigl\| H_j - \bar{H}_{\text{group}}^{(g)} \bigr\|_{\mathrm{F}}^2 \;\le\; \sum_{j \in \mathcal{G}_g} \bigl\| H_j - \bar{H}_{\text{global}} \bigr\|_{\mathrm{F}}^2
\end{equation}

\noindent\textit{Remark.} The maximum per-block deviation $\max_{j}\|H_j - \bar{H}_{\text{group}}^{(g)}\|_{\mathrm{F}}$ need not be smaller than $\|H_j - \bar{H}_{\text{global}}\|_{\mathrm{F}}$ for the same~$j$; Proposition~1 is the correct ``variance reduction'' statement. Translating Lemma~1 and Proposition~1 into a bound on the final attention \emph{output} requires tracking softmax normalization and the zeroth-order term; we omit that bookkeeping here and treat Lemma~1 as a direct handle on the first-order matrix residual.







\section{Online Resources}
Our code is available at \url{https://anonymous.4open.science/r/D28D/}.

\section{visualization of the generated videos}
\label{sec:visualization-generated-videos}

For qualitative visualization, we randomly select 5 videos generated by Wan2.1-T2V-1.3B; each sample is a 5-second clip at 480p resolution. Due to file size limits, the displayed images are compressed.

\begin{figure*}[!t]
  \centering
  \begin{subfigure}[b]{0.9\textwidth}
    \centering
    \includegraphics[height=0.14\textheight,keepaspectratio]{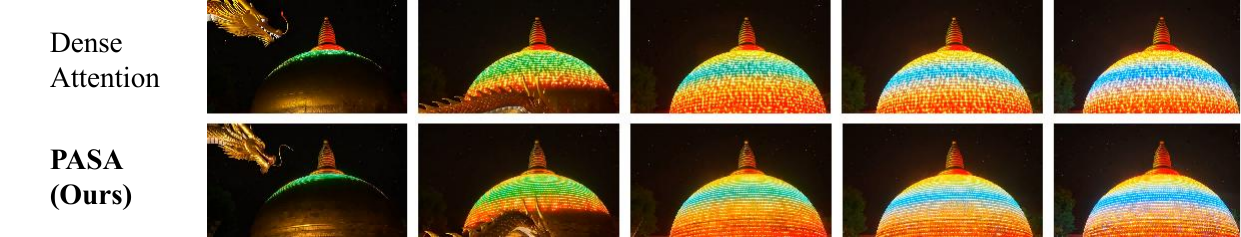}
    \caption{Visualization for dense attention and our method with prompt \textit{A benevolent dragon circles above an ancient temple as dawn light illuminates the sky, and the camera slowly zooms in on this dreamlike scene.}}
  \end{subfigure}
  \par\medskip
  \begin{subfigure}[b]{0.9\textwidth}
    \centering
    \includegraphics[height=0.14\textheight,keepaspectratio]{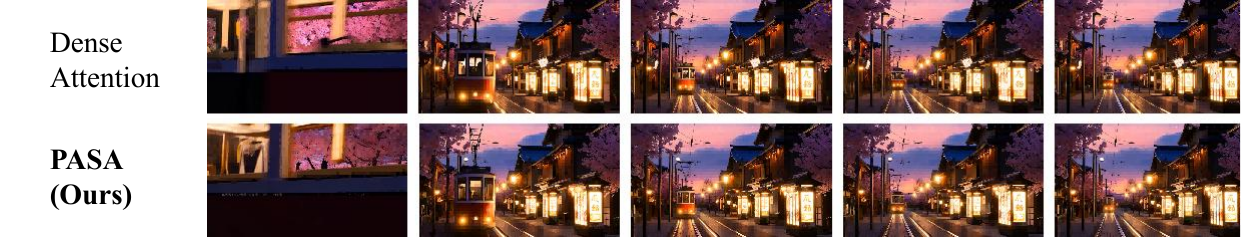}
    \caption{Visualization for dense attention and our method with prompt \textit{The camera pans through cherry blossoms to a Japanese street, then follows a tram under a pink-blue sky with warm evening lights.}}
  \end{subfigure}
  \par\medskip
  \begin{subfigure}[b]{0.9\textwidth}
    \centering
    \includegraphics[height=0.14\textheight,keepaspectratio]{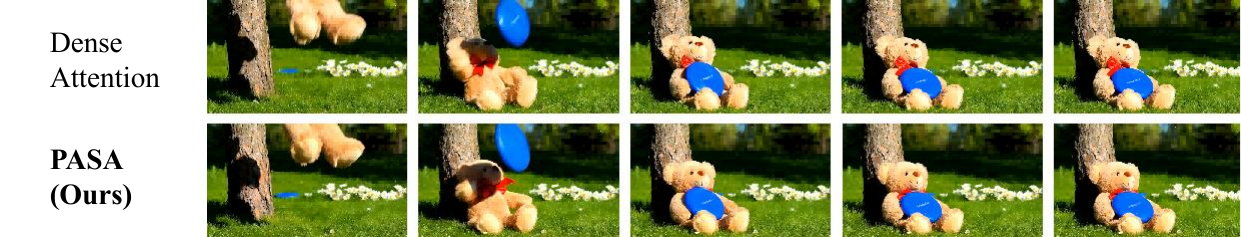}
    \caption{Visualization for dense attention and our method with prompt \textit{A plush teddy bear and a blue frisbee play on a sunny lawn, ending with the bear resting by a tree in a warm, playful scene.}}
  \end{subfigure}
  \par\medskip
  \begin{subfigure}[b]{0.9\textwidth}
    \centering
    \includegraphics[height=0.14\textheight,keepaspectratio]{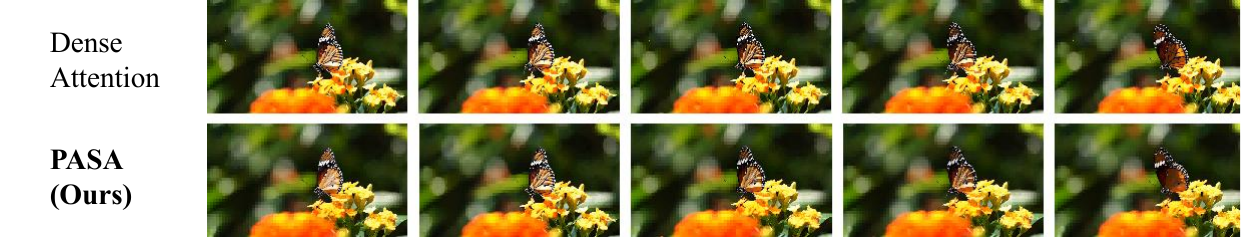}
    \caption{Visualization for dense attention and our method with prompt \textit{A butterfly is fluttering.}}
  \end{subfigure}
  \par\medskip
  \begin{subfigure}[b]{0.9\textwidth}
    \centering
    \includegraphics[height=0.14\textheight,keepaspectratio]{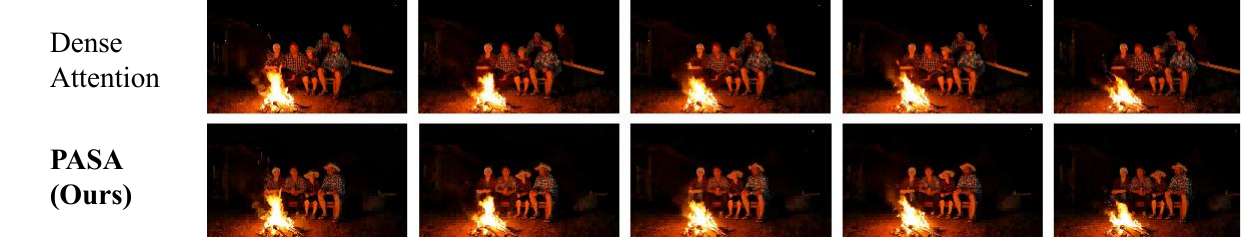}
    \caption{Visualization for dense attention and our method with prompt \textit{At the entrance of the village at night, five farmers are warming themselves by the fire.}}
  \end{subfigure}
  \caption{Visualization results of the generated videos.}
  \label{fig:generated-video-visualization}
\end{figure*}

\end{document}